# Leveraging Large Language Models for Web Scraping

*Aman Ahluwalia, PerpetualBlock Technologies Pvt. Ltd, Pune, Maharashtra*
*Suhrud Wani, Innoplexus Consulting Services Pvt. Ltd., Pune, Maharashtra*

**ABSTRACT**: Large Language Models (LLMs) demonstrate remarkable capabilities in replicating human tasks and boosting productivity. However, their direct application for data extraction presents limitations due to a prioritization of fluency over factual accuracy and a restricted ability to manipulate specific information. Therefore to overcome these limitations, this research leverages the knowledge representation power of pre-trained LLMs and the targeted information access enabled by RAG models, this research investigates a general-purpose accurate data scraping recipe for RAG models designed for language generation.

To capture knowledge in a more modular and interpretable way, we use pre trained language models with a latent knowledge retriever, which allows the model to retrieve and attend over documents from a large corpus. We utilised RAG model architecture and did an in-depth analysis of their capabilities under three tasks: (i) Semantic Classification of HTML elements, (ii) Chunking HTML text for effective understanding, and (iii) comparing results from different LLMs and ranking algorithms. While previous work has developed dedicated architectures and training procedures for HTML understanding and extraction, we show that LLMs pretrained on standard natural language with an addition of effective chunking, searching and ranking algorithms, can prove to be efficient data scraping tool to extract complex data from unstructured text.

Future research directions include addressing the challenges of provenance tracking and dynamic knowledge updates within the proposed RAG-based data extraction framework. By overcoming these limitations, this approach holds the potential to revolutionize data extraction from vast repositories of textual information.

**KEYWORDS:** Large Language Models, Data Scraping , RAG, Vector Embeddings, Text Chunking, Vector Stores

## I. INTRODUCTION

The World Wide Web is a vast repository of information, with billions of active web pages constantly evolving and presenting both opportunities and challenges for data extraction. Traditional web crawling methods, reliant on predefined rules, struggle to keep pace with the dynamic nature of the web, often missing relevant content. In this paper, we explore how Large Language Models (LLMs) coupled with prompt engineering can revolutionize web crawling by enabling simpler, faster, and more accurate data extraction.

Traditional web crawling methods face limitations in areas such as dynamic content, semantic understanding, and evolving language. Factors like the increasing use of JavaScript frameworks, the difficulty in grasping context and meaning, and the ever-changing nature of language hinder the effectiveness of traditional approaches. While web crawling offers valuable insights and data for various purposes such as research, business intelligence, and data analysis, it's crucial to ensure that such methods are deployed responsibly and ethically. This entails respecting website terms of service, privacy policies, and robots.txt directives to avoid violating the rights of website owners and users.

LLMs, trained on massive datasets of text and code, possess exceptional abilities in understanding language nuances and context. Coupled with prompt engineering, they can perform targeted tasks like data extraction with enhanced semantic understanding, intelligent link extraction, and adaptation to dynamic content. This enables them to extract information more accurately and comprehensively compared to traditional methods.

Studies have shown that LLM-based crawling reduces the time needed to collect data and increases data accuracy compared to traditional methods. Market research shows LLM-based crawling reduced data collection time by 25% while increasing data accuracy by 10%. Similarly, in e-commerce, LLMs achieved 92% precision in extracting product information compared to 85% achieved by traditional methods

## II. Retrieval Augmented Generation

Large Language Models (LLMs) have changed many fields by being really good at things like writing, translating, and answering questions. But they have limits because they only know what's inside their own memory. This makes it hard for them to do tasks that need real-world knowledge or understanding. To fix this problem, Retrieval Augmented Generation (RAG) was created.

Challenges of Traditional LLM-based Data Extraction:

1. Incomplete Knowledge Coverage: LLMs are trained on massive amounts of text data, but this data may not encompass the entirety of human knowledge or the specific context of a given extraction task.

This can lead to the extraction of inaccurate or irrelevant information.
2. Struggle with Implicit Information: LLMs often struggle with implicit information, such as relationships between concepts or specific domain knowledge, hindering their ability to extract the full depth and meaning from web pages.
3. Dynamic Content Limitations: Websites increasingly utilize dynamic content generated by JavaScript, which can be invisible to traditional LLM-based crawlers. This limits their ability to access and extract valuable data.
4. Hallucination: LLMs can generate information that appears plausible but is entirely fabricated or incorrect, a phenomenon known as hallucination. This can result in the extraction of non-existent data or misleading details, compromising the reliability of the extracted information.

In the realm of Retrieval Augmented Generation (RAG), Large Language Models (LLMs) play a pivotal role in enhancing the accuracy, consistency, completeness, reliability, and relevance of data extraction and validation processes. LLMs, such as GPT-4, are proficient in understanding and generating human-like text based on vast amounts of training data. Leveraging this capability, RAG combines retrieval-based techniques with generative models to retrieve relevant information from large datasets and generate coherent responses. By incorporating retrieval mechanisms, LLMs can access a wealth of contextual information, thereby improving the accuracy and relevance of extracted data. Moreover, LLMs contribute to consistency by ensuring uniformity in language usage and formatting across extracted content. Furthermore, their ability to generate detailed and comprehensive responses enhances the completeness of extracted data. In terms of reliability, LLMs offer a consistent performance in extracting and validating data, reducing the likelihood of errors or inconsistencies. Overall, the synergy between LLMs and RAG techniques elevates the efficacy of data extraction and validation processes, fostering greater trust in the reliability and quality of extracted information.

**Advantages of RAG**:
1. It improves accuracy by using external knowledge to make sure the extracted data is accurate and complete.
2. It also enhances understanding by helping understand implied information, making the data extraction more comprehensive.
3. Can deal with dynamic content, extracting data that traditional methods might miss.

**Text Chunking**
RAG models inherently face limitations due to the restricted context window of Large Language Models (LLMs). Feeding an entire document at once would overwhelm the LLM and hinder its ability to identify relevant information. So we break down the text into manageable segments, each containing a coherent unit of information. By presenting the LLM with focused chunks, the retrieval process becomes more targeted, leading to a higher likelihood of finding relevant passages for response generation.

Chunking reduces the computational burden on the LLM by limiting the amount of text it needs to process at once. This translates to faster response generation and avoids overwhelming the model with irrelevant details, ultimately improving the accuracy and focus of the response. We tailored our chunking strategy to maintain the inherent structure and flow of the original text. This is particularly important for tasks like document summarization or question answering, where preserving context is crucial for generating coherent and informative responses.

For more tailored chunking, we utilize the Recursive Character Text Splitting (RCTS) technique within our LLM-based data extraction pipeline. RCTS facilitates efficient processing and extraction by iteratively splitting the web page content into smaller, manageable chunks.

Recursive Character Text Splitting (RCTS) is a powerful method for splitting text into manageable chunks for various NLP tasks, particularly data extraction with LLMs. This document delves into the technical details of the RCTS algorithm, explaining its functionalities and its relevance within the our proposed framework.

**Algorithm Breakdown:**
The RCTS operates on a core principle: recursively attempting to split the text using a predefined list of characters until the resulting chunks fall below a specified size limit. This splitting is trying to keep related pieces of text next to each other. This is the recommended way to start splitting text.

The process of recursively splitting text begins with an initial input consisting of a raw text string, a list of characters used for splitting (ordered from the largest delimiter, such as "\n\n", to the smallest, such as a space), and a specified maximum chunk size limit. The algorithm iterates through the list of delimiters, splitting the text at each occurrence of the current delimiter, which results in potential sub-texts. Each sub-text is then evaluated: if its length is within the chunk size limit, it is added to the final list of chunks. If a sub-text exceeds the limit, the function is called recursively on that sub-text using the remaining delimiters. This recursive process continues until all delimiters have been exhausted for a particular sub-text. If any remaining sub-text still exceeds the chunk size limit at this stage, it is added to the final list as the largest possible chunk. Ultimately, the function returns a list of text chunks that conform to the specified size constraints.

**Benefits and Considerations :**
- **Context Preservation**: By attempting splits based on larger delimiters first (e.g., paragraphs, sentences), RCTS prioritizes preserving contextual relationships within the text. This is crucial for tasks like data extraction, where maintaining context is essential for accurate information retrieval.
- **Flexibility**: The customizable char_list allows users to tailor the splitting behavior to their specific needs. For instance, including punctuation in the list can create smaller chunks suitable for tokenization tasks.
- **Efficiency**: The recursive approach ensures that splitting occurs only when necessary, avoiding unnecessary processing for already small text segments.

However, it's important to consider limitations:
- **Computational Cost**: Deep recursion can lead to increased processing time for very large documents.
- **Over-splitting**: Depending on the char_list configuration, RCTS might create excessively small chunks for certain types of text, potentially harming tasks that rely on larger context windows.

**Vector Stores**

At the heart of our architecture lies the concept of vector embeddings. These embeddings represent textual data as dense numerical vectors, where similar documents or concepts map to vectors with closer proximity in the vector space. This allows for efficient similarity searches, enabling the retrieval of documents most relevant to a given query. With various vector stores, including popular options like Pinecone and Faiss, offering developers a flexible choice based on specific project requirements.

**Key Concepts for Data Extraction**
- **Vector Indexes**: They facilitate the creation and management of vector indexes within a chosen vector store. It allows developers to define the configuration for embedding data and enables efficient indexing for subsequent retrieval operations.
- **Document**: It can hold not only the raw text data but also its corresponding vector embedding, facilitating the association of textual content with its semantic representation in the vector space.
- **Text Embedding Model Interfaces**: These models convert textual data from documents into vector representations, allowing for integration with different embedding techniques based on project needs.
- Similarity Search Functionality: This enables the retrieval of documents with embeddings closest to the query vector, thereby identifying the most relevant data points for extraction.

**Workflow for Data Extraction using Vector Stores**

Here's a breakdown of the workflow for data extraction using vector libraries :
1. **Data Preprocessing**: Textual data is preprocessed and cleaned to ensure optimal performance of the embedding model. This may involve tasks like tokenization, stop word removal, and stemming/lemmatization.
2. **Text Embedding**: The preprocessed text is fed into the chosen text embedding model, generating a dense vector representation for each document.
3. **Vector Indexing**: The document objects, containing both the text and its corresponding vector embedding, are indexed within the chosen vector store using the VectorstoreIndexCreator.
4. **Query Embedding**: The user query is also processed and converted into a vector representation using the same text embedding model.
5. **Similarity Search**: The query vector is used to search the indexed vectors within the vector store. Documents with embeddings closest to the query vector are retrieved based on a predefined similarity metric.
6. **LLM Processing**: The retrieved documents, containing the most relevant information based on semantic similarity, are fed into the LLM. The LLM can then process this contextually rich data to extract the desired information with greater accuracy and efficiency.

By leveraging vector libraries, we can overcome the limitations of LLMs for direct data extraction. This approach facilitates the retrieval of highly relevant information based on semantic similarity, ultimately leading to more accurate and efficient data extraction pipelines.

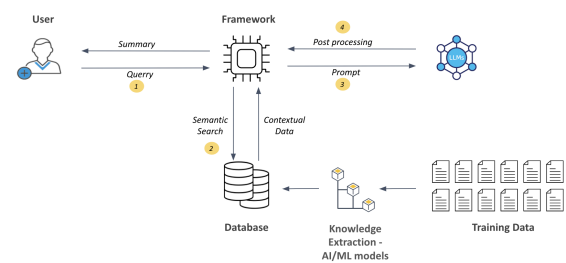

### III. Methodology

The first step involves preparing the data from which information will be scraped. This stage begins with a collection of website URLs that serve as the target source for data extraction. To enable the LLM to process this data effectively, these URLs are subsequently converted into HTML content pages through browser rendering. This rendering process fetches the complete HTML code of the webpage, encompassing all the structured and unstructured data it contains. This retrieved HTML serves as the raw material for the LLM to analyze and extract the desired information.

To enhance the search process and facilitate efficient information retrieval, the retrieved HTML content is further divided into smaller, manageable chunks. This chunking process aims to break down the webpage into more granular components that can be efficiently processed by the LLM. Each chunk may represent a specific section of the required data we will be fetching in the next steps.

Following the chunking process, these data segments undergo an embedding process. Text embedding techniques convert the textual data within each chunk into a higher-dimensional numerical representation. This compressed format allows for more efficient similarity comparisons between the chunked data and the user's scraping query. There are many embedding methods which map words to vectors based on their context and co-occurrence within the text. By embedding the chunked data, the system creates a searchable index that the LLM can leverage to identify relevant sections containing the target information.

One of the most common ways to store and search over unstructured data is to embed it and store the resulting embedding vectors, and then at query time to embed the query and retrieve the embedding vectors that are 'most similar' to the embedded query. A vector store takes care of storing embedded data and performing vector search for you. Embeddings create a vector representation of a piece of text which is useful because it means we can think about text in the vector space, and do tasks like semantic search where we look for pieces of text that are most similar in the vector space.

These embedding vectors are subsequently stored in a specialized data structure called a FAISS (Facebook AI Similarity Search) vector store. This storage mechanism is optimized for efficient retrieval based on vector similarity. When a query to yield a particular field is submitted to the system , it is also converted into an embedding vector. The system then performs a similarity search within the FAISS vector store, identifying the top k chunks (where k is a predefined value) that exhibit the highest degree of semantic similarity to the query vector.

The retrieved top k chunks, containing the HTML content exhibiting the highest semantic alignment with the presence of a particular field we are looking to extract, are then concatenated. This process essentially combines these relevant chunks into a single, cohesive unit for the LLM to analyze.

Next, carefully crafted prompts (each tailored to yield a particular field value from the HTML) are employed to guide the LLM in extracting the specific data points of interest. These prompts act as instructions, providing context and specifying the desired information within the concatenated chunks. The LLM's capability to understand and respond to natural language allows it to interpret these prompts and navigate the retrieved HTML content. By effectively combining the relevant chunks and incorporating informative prompts, the system unlocks the LLM's potential to pinpoint and extract the precise data sought by the user.

To address potential hallucinations introduced by individual large language models (LLMs), we employed an ensemble approach utilizing three distinct LLMs. This strategy leverages the strengths of each LLM while mitigating potential biases or errors.

**Mixtral AI:** This LLM was chosen due to its demonstrated performance in retrieval-based tasks. It offers competitive capabilities at an advantageous cost-performance ratio. Additionally, its multilingual support (English, French, Spanish, German, Italian) allows for broader applicability.

**OpenAI GPT-4.0:** This well-established LLM excels at various natural language processing tasks, including information extraction. We leveraged its ability to generate normalized data outputs that are well-suited for database integration through the application of data template constraints within the prompts.

**Llama 3:** This LLM combines retrieval and generation capabilities. It was included for its versatility in handling scenarios where external data integration is required while maintaining informative and coherent text generation.

**Ensemble Voting for Enhanced Accuracy**
Following the independent processing of data by each LLM, we implemented a ranking algorithm based on an ensemble voting mechanism. Each LLM evaluated the outputs from all three models, considering factors like accuracy, data frequency within the outputs, and overall data quality. "Accuracy" was determined by comparing the extracted data against a predefined ground truth dataset. "Data frequency" referred to the consistency of a particular data point across the LLM outputs. "Data quality" assessed the completeness and internal consistency of the extracted information. The final result was designated as the option receiving the most votes from the ensemble.

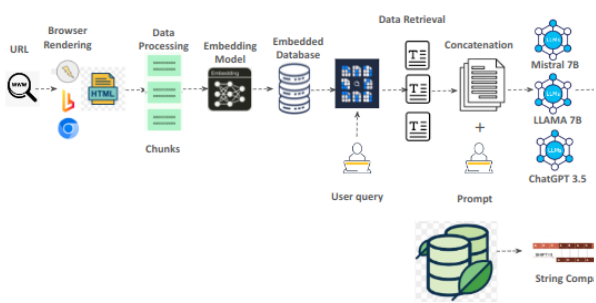

## IV. Future Work

Future research directions hold immense potential to further refine our RAG-based data extraction framework. A crucial area of exploration is the concept of fine-tuning the LLMs employed within the system. LLMs, while powerful, can exhibit biases or knowledge gaps that hinder their performance in specific domains. Fine-tuning these models with domain-specific data can significantly enhance their ability to grasp the intricacies of relevant terminology and website structures. This can be achieved by training the LLMs on datasets tailored to the target extraction task. For instance, if the goal is to extract product information from e-commerce websites, fine-tuning the LLMs with product descriptions, specifications, and pricing data can significantly improve their ability to identify and extract relevant information from product pages. By incorporating domain-specific knowledge, fine-tuned LLMs can become more adept at understanding the nuances of the target websites, leading to more accurate, comprehensive, and relevant data extraction. This targeted approach can pave the way for the development of specialized RAG models tailored to various web scraping applications.

## V. References

(1) Dhruv Patel , Shrey Pandey, Abhishek Sharma et al. Efficient Vector Store System for Python using Shared Memory Association. Association of Computing Machinery (2023).
(2) Luc Patiny, Guillaume Godin et al. Automatic extraction of FAIR data from publications using LLM. ChemRxiv (2023)
(3) Patrick Lewis, Ethan Perez, Aleksandra Piktus, Fabio Petroni, Vladimir Karpukhin, Naman Goyal, Heinrich Küttler, Mike Lewis, Wen-tau Yih, Tim Rocktäschel, Sebastian Riedel, Douwe Kiela. Retrieval-Augmented Generation for Knowledge-Intensive NLP Tasks. 34th Conference on Neural Information Processing Systems (NeurIPS 2020), Vancouver, Canada (2020)
(4) Izzeddin Gur, Ofir Nachum, Yingjie Miao, Mustafa Safdari, Austin Huang, Aakanksha Chowdhery, Sharan Narang, Noah Fiedel, Aleksandra Faust . UNDERSTANDING HTML WITH LARGE LANGUAGE MODELS .
(5) Zhang Yao ,Wang Daling ,Feng Shi , Zhang Yifei , Leng Fangling An Approach for Crawling Dynamic WebPages Based on Script Language Analysis. 2012 Ninth Web Information Systems and Applications Conference. (2012)
(6) Kelvin Guu, Kenton Lee, Zora Tung, Panupong Pasupat, Ming-Wei Chang . REALM: Retrieval-Augmented Language Model Pre-Training .
(7) Goel, A., Gueta, A., Gilon, O., Liu, C., Erell, S., Nguyen, L.H., Hao, X., Jaber, B., Reddy, S., Kartha, R., Steiner, J., Laish, I. & Feder, A.. (2023). LLMs Accelerate Annotation for Medical Information Extraction. <i>Proceedings of the 3rd Machine Learning for Health Symposium</i>, in <i>Proceedings of Machine Learning Research</i> 225:82-100
(8) Armen Aghajanyan, Dmytro Okhonko, Mike Lewis, Mandar Joshi, Hu Xu, Gargi Ghosh, and Luke Zettlemoyer. Html: Hyper-text pre-training and prompting of language models. arXiv preprint arXiv:2107.06955, 2021.